\documentclass{article}
\usepackage{kr}

\usepackage{times}
\usepackage{soul}
\usepackage{url}
\usepackage[hidelinks]{hyperref}
\usepackage[utf8]{inputenc}
\usepackage[small]{caption}
\usepackage{graphicx}
\usepackage{amsmath}
\usepackage{amsthm}
\usepackage{latexsym}
\usepackage{amssymb}
\usepackage{nccmath}

\usepackage{booktabs}
\usepackage{algorithm}
\usepackage{algorithmicx}
\usepackage[noend]{algpseudocode}
\usepackage{multirow}
\usepackage{verbatim}
\usepackage{titlesec}
\usepackage{setspace}
\usepackage{enumitem}
\usepackage{subcaption}
\usepackage{relsize}

\usepackage{adjustbox}         
\usepackage[table]{xcolor}
\usepackage{multirow,siunitx}

\usepackage{color}
\usepackage{url}
\usepackage{stmaryrd}
\usepackage{soul}

\usepackage{pgf}
\usepackage{tikz}
\usepackage{forest}

\usetikzlibrary{arrows,decorations.pathmorphing,decorations.footprints,fadings,calc,trees,mindmap,shadows,decorations.text,patterns,positioning,shapes,matrix,fit}
\usetikzlibrary{arrows,shadows,backgrounds}
\usetikzlibrary{arrows.meta}
\usetikzlibrary{positioning,fit}
\usetikzlibrary{automata}
\usetikzlibrary{matrix}
\usetikzlibrary{shapes.symbols,shapes.misc,shapes.arrows}
\usetikzlibrary{matrix,chains,scopes,decorations.pathmorphing}
\usetikzlibrary{bayesnet}

\makeatletter
\newtheorem*{rep@theorem}{\rep@title}
\newcommand{\newreptheorem}[2]{%
\newenvironment{rep#1}[1]{%
 \def\rep@title{#2 \ref{##1}}%
 \begin{rep@theorem}}%
 {\end{rep@theorem}}}
\makeatother

\newtheorem{proposition}{Proposition}

\newtheorem{definition}{Definition}

\newtheorem{example}{Example}

\newreptheorem{theorem}{Theorem}
\newreptheorem{proposition}{Proposition}
\newreptheorem{lemma}{Lemma}

\newenvironment{Proof}{\noindent{\em Proof.~}}{\hfill$\Box$\\[-0.25cm]}

\newcommand{\todoF}[2]{}

\newcommand{\fml}[1]{{\mathcal{#1}}}

\newcommand{\tn}[1]{\textnormal{#1}}

\newcommand{\mbf}[1]{\ensuremath\mathbf{#1}}
\newcommand{\mbb}[1]{\ensuremath\mathbb{#1}}

\newcommand{\mfrak}[1]{\ensuremath\mathfrak{#1}}

\newcommand{\childs}{\mathsf{children}}
\newcommand{\parent}{\mathsf{parent}}
\newcommand{\isterm}{\mathsf{isTerminal}}

\newcommand{\rootG}{\mathsf{root}}
\newcommand{\emptyQ}{\mathsf{empty}}
\newcommand{\initQ}{\mathsf{init}}
\newcommand{\addQ}{\mathsf{enqueue}}
\newcommand{\delQ}{\mathsf{dequeue}}

\newcommand{\axp}{\ensuremath\mathsf{findAXp}}
\newcommand{\cxp}{\ensuremath\mathsf{findCXp}}

\newcommand{\cpath}{\ensuremath\mathsf{pathToZero}}

\newcommand{\SAT}{\ensuremath\mathsf{SAT}}
\newcommand{\outc}{\ensuremath\mathsf{outc}}
\newcommand{\prtaxp}{\ensuremath\mathsf{reportAXp}}
\newcommand{\prtcxp}{\ensuremath\mathsf{reportCXp}}

\newcommand{\stwop}{\Sigma_2^\tn{P}}

\DeclareMathOperator*{\lequiv}{\leftrightarrow}
\DeclareMathOperator*{\limply}{\rightarrow}

\DeclareMathOperator*{\outdeg}{\tn{deg}^{\tn{+}}}

\definecolor{gray}{rgb}{.4,.4,.4}
\definecolor{midgrey}{rgb}{0.5,0.5,0.5}
\definecolor{middarkgrey}{rgb}{0.35,0.35,0.35}
\definecolor{darkgrey}{rgb}{0.3,0.3,0.3}
\definecolor{midred}{rgb}{0.7,0.2,0.2}
\definecolor{darkred}{rgb}{0.7,0.1,0.1}
\definecolor{midblue}{rgb}{0.2,0.2,0.7}
\definecolor{darkblue}{rgb}{0.1,0.1,0.5}
\definecolor{midgreen}{rgb}{0.2,0.5,0.2}
\definecolor{darkgreen}{rgb}{0.1,0.5,0.1}
\definecolor{defseagreen}{cmyk}{0.69,0,0.50,0}

\newcommand{\jnoteF}[1]{}

\algnewcommand\algorithmicinput{\textbf{input:}}
\algnewcommand\Input{\item[\algorithmicinput]}
\newcommand{\True}{\textbf{true}}
\newcommand{\False}{\textbf{false}}
\newcommand{\NOT}{\textbf{not}}

\newcommand{\OR}{\textbf{or}}
\newcommand{\RETURN}{\textbf{return}}

\tikzset{
  0 my edge/.style={densely dashed, my edge},
  my edge/.style={-{Stealth[]},thin},
}

\newcolumntype{L}[1]{>{\raggedright\let\newline\\\arraybackslash\hspace{0pt}}m{#1}}
\newcolumntype{C}[1]{>{\centering\let\newline\\\arraybackslash\hspace{0pt}}m{#1}}
\newcolumntype{R}[1]{>{\raggedleft\let\newline\\\arraybackslash\hspace{0pt}}m{#1}}

\title{On Efficiently Explaining Graph-Based Classifiers}

\author{%
Xuanxiang Huang$^1$\and
Yacine Izza$^1$\and
Alexey Ignatiev$^2$\and
Joao Marques-Silva$^3$ \\
\affiliations
$^1$University of Toulouse, France\\
$^2$Monash University, Melbourne, Australia\\
$^3$IRIT, CNRS, Toulouse, France\\
\emails
\{xuanxiang.huang,yacine.izza\}@univ-toulouse.fr,
alexey.ignatiev@monash.edu,
joao.marques-silva@irit.fr
}

\begin{document}

\maketitle

\begin{abstract}
  Recent work has shown that not only decision trees (DTs) may not be 
  interpretable but also proposed a polynomial-time algorithm for 
  computing one PI-explanation of a DT. 
  This paper shows that for a wide range of classifiers, globally
  referred to as decision graphs, and which include decision trees and 
  binary decision diagrams, but also their multi-valued variants,
  there exist polynomial-time algorithms for computing one
  PI-explanation. In addition, the paper also proposes a
  polynomial-time algorithm for computing one contrastive
  explanation.
  These novel algorithms build on explanation graphs (XpG's). XpG's
  denote a graph representation that enables both theoretical and
  practically efficient computation of explanations for decision
  graphs.
  Furthermore, the paper proposes a practically efficient solution for
  the enumeration of explanations, and studies the complexity of
  deciding whether a given feature is included in some explanation. For the concrete case
  of decision trees, the paper shows that the set of all contrastive
  explanations can be enumerated in polynomial time.
  Finally, the experimental results validate the practical
  applicability of the algorithms proposed in the paper on a wide
  range of publicly available benchmarks.
\end{abstract}

\section{Introduction} \label{sec:intro}

The emerging societal impact of Machine Learning (ML) and its foreseen
deployment in safety critical applications, puts additional demands
on approaches for verifying and explaining ML
models~\cite{weld-cacm19}.
The vast majority of approaches for explainability in ML (often
referred to as eXplainable AI (XAI)~\cite{darpa-xai}) are heuristic,
offering no formal guarantees of soundness, with well-known examples
including tools like LIME, SHAP or
Anchors~\cite{guestrin-kdd16,lundberg-nips17,guestrin-aaai18}.
(Recent surveys~\cite{pedreschi-acmcs19} cover a wider range of
heuristic methods.)
Moreover, recent work has shed light on the important practical
limitations of heuristic XAI
approaches~\cite{nsmims-sat19,inms-corr19,lukasiewicz-corr19,lakkaraju-aies20a,lakkaraju-aies20b,ignatiev-ijcai20}.

In contrast, formal approaches to XAI have been proposed in recent
years~\cite{darwiche-ijcai18,inms-aaai19,darwiche-aaai19,inms-nips19,darwiche-ecai20,marquis-kr20}
(albeit it can be related to past work on logic-based explanations
(e.g.\ \cite{shanahan-ijcai89,simari-aij02,uzcategui-aij03})).
The most widely studied form of explanation consists in the
identification of prime implicants (PI) of the decision function
associated with an ML classifier, being referred to as
PI-explanations.
Although PI-explanations offer important formal guarantees, e.g.\ they
represent minimal sufficient reasons for a prediction, they do have 
their own drawbacks.
First, in most settings, finding one PI-explanation is NP-hard, and in
some settings scalability is an
issue~\cite{darwiche-ijcai18,inms-aaai19}. Second, users have little
control on the size of computed PI-explanations (and it is well-known
the difficulty that humans have in grasping complex concepts).
Third, there can be many PI-explanations, and it is often unclear
which ones are preferred. Fourth, in practice users may often prefer
high-level explanations, in contrast with feature-based, low-level
explanations.
Despite these drawbacks, it is plain that PI-explanations offer a
sound basis upon which one can expect to develop theoretically sound
and practically effective approaches for computing explanations.
For example, more recent work
has demonstrated the tractability of PI-explanations for some ML
models~\cite{marquis-kr20,msgcin-nips20,msgcin-icml21,iincms-corr21},
in some cases allowing for polynomial delay
enumeration~\cite{msgcin-nips20}.
Also, recent
work~\cite{ignatiev-ijcai20,ims-ijcai21,ims-sat21,iincms-corr21}
showed that, even for ML models for which computing a PI-explanation
is NP-hard, scalability may not be an obstacle.

Moreover, it was recently shown
that finding explanations can be crucial even for ML models that are
generally deemed interpretable\footnote{%
  Interpretability is regarded a subjective concept, with no accepted
  rigorous definition~\cite{lipton-cacm18}. 
  In this paper,
  we equate interpretability with explanation succinctness.
}.
One such example are decision trees~\cite{iims-corr20}. 
Decision trees (DTs) are not only among the most widely used 
ML models, but are also generally regarded as interpretable~\cite{breiman-ss01,freitas-sigkdd13,guestrin-corr16,muller-dsp18,muller-bk19,molnar-bk19,miller-aij19,pedreschi-acmcs19,rudin-naturemi19,zhu-nlpcc19,gombolay-aistats20}.
However, recent work~\cite{iims-corr20} has shown that paths in DTs
may contain literals that are irrelevant for identifying minimal
sufficient reasons for a prediction, and that the number of redundant
literals can grow asymptotically as large as the number of features. 
Furthermore, it was also shown~\cite{iims-corr20} that PI-explanations
for DTs can be computed in polynomial time.
Moreover, independent work showed that finding a smallest explanation
is hard for NP~\cite{barcelo-nips20}, thus hinting at the need to
finding PI-explanations in the case of DTs.

This paper complements this earlier work with several novel results.
First, the paper considers both PI (or abductive) explanations (AXps)
and contrastive explanations
(CXps)~\cite{miller-aij19,inams-aiia20}, which will be jointly
referred to as explanations (XPs). 
Second, the paper shows that XPs can be computed in polynomial time
for a much larger class of classifiers, which will be conjointly
referred to as \emph{decision
  graphs}~\cite{oliver-tr92,kohavi-aaai94}%
\footnote{%
  The term \emph{decision graph} is also 
  used in the context of Bayesian
  Networks~\cite{jensen-bk01,darwiche-bk09}, and more recently in
  explainability~\cite{darwiche-ijcai18,darwiche-aaai19}.
  However, and to the best of our knowledge, the term ``decision
  graph'' was first proposed 
  in the early 90s~\cite{oliver-tr92} to enable more compact
  representation of DTs.
}.
For that, the paper introduces a new graph representation, namely the
\emph{explanation graph}, and shows that for any classifier (and
instance) that can be reduced to an explanation graph, XPs can be
computed in polynomial time. (For example, multi-valued variants of
decision trees, graphs or diagrams can be reduced to explanation
graphs.)
The paper also shows that the MARCO algorithm for enumerating
MUSes/MCSes~\cite{lpmms-cj16} can be adapted to the enumeration of
XPs, yielding a solution that is very efficient in practice.
For the case of DTs, the paper proves that the set of all CXps
can be computed in polynomial time. In turn, this result offers an
alternative approach for the enumeration of PI-explanations,
e.g.\ based on hitting set
dualization~\cite{reiter-aij87,liffiton-jar08}.
Finally, we investigate the \emph{explanation membership
  problem}, i.e.\ to decide whether a feature (given its assigned
value) can be included in some explanation (either AXp or CXp). The
paper shows that for arbitrary explanation graphs, the explanation
membership problem is in $\tn{NP}$, while for a propositional formula
in disjunctive normal form (DNF) is shown to be hard for $\stwop$.
However, for tree explanation graphs (which can represent explanations
of decision trees), deciding explanation membership is shown to be in
$\tn{P}$. 

The paper is organized as follows.
\autoref{sec:prelim} introduces the definitions and notation used in
the rest of the paper.
\autoref{sec:xpg} studies explanation graphs (XpG's), and shows how
XpG's can be used for computing explanations.
Afterwards, \autoref{sec:xps} describes algorithms computing one XP
(either AXp or CXp) of XpG's, and a MARCO-like algorithm for the
enumeration of XPs. \autoref{sec:xps} also proves that for DTs,
the set of all CXps can be computed in polynomial time.
Some of the previous results are used in~\autoref{sec:chkxp} for
investigating the complexity of deciding membership of features in
explanations.
\autoref{sec:relw} relates the paper's contributions with earlier
work.
\autoref{sec:res} discusses experimental results of explaining DTs and
reduced ordered binary decision diagrams, including AXps, CXps and
their enumeration.
Finally, the paper concludes in~\autoref{sec:conc}.

\section{Preliminaries} \label{sec:prelim}

\paragraph{Classification problems.}
A classification problem is defined on a set of features (or
attributes) $\fml{F}=\{1,\ldots,m\}$ and a set of classes
$\fml{K}=\{c_1,c_2,\ldots,c_K\}$.
Each feature $i\in\fml{F}$ takes values from a domain $\mbb{D}_i$.
Domains can be boolean, integer or real-valued.
Feature space is defined as
$\mbb{F}=\mbb{D}_1\times{\mbb{D}_2}\times\ldots\times{\mbb{D}_m}$.
The notation $\mbf{x}=(x_1,\ldots,x_m)$ denotes an arbitrary point in
feature space, where each $x_i$ is a variable taking values from
$\mbb{D}_i$. Moreover, the notation $\mbf{v}=(v_1,\ldots,v_m)$
represents a specific point in feature space, where each $v_i$ is a
constant representing one concrete value from $\mbb{D}_i$.
An \emph{instance} (or example) denotes a pair $(\mbf{v}, c)$, where
$\mbf{v}\in\mbb{F}$ and $c\in\fml{K}$. (We also use the term
\emph{instance} to refer to $\mbf{v}$, leaving $c$ implicit.)
An ML classifier $\mbb{C}$ is characterized by a \emph{classification
function} $\kappa$ that maps feature space $\mbb{F}$ into the set of
classes $\fml{K}$, i.e.\ $\kappa:\mbb{F}\to\fml{K}$. ($\kappa$ is
assumed to be non-constant.)

\paragraph{Remark on binarization.}
We underline the importance of not restricting feature domains to be
boolean-valued.
Although binarization can be used to represent features that are
categorical, integer or real-valued, it is also the case that, from
the perspective of computing explanations, soundness demands that one
must know whether binarization was applied and, if so, which resulting
binary features must be related with which original features. The key
observation is that if binarization is used, then soundness of results
imposes that features \emph{must} be reasoned about in groups
of related binary features, and this implies algorithms that work
under this assumption.
In this paper, we opt to impose no such restriction when reasoning
about explanations.

\paragraph{Abductive and constrastive explanations.}
We now define formal explanations.
Prime implicant (PI) explanations~\cite{darwiche-ijcai18} denote a
minimal set of literals (relating a feature value $x_i$ and a constant
$v_i\in\mbb{D}_i$ 
that are sufficient for the prediction\footnote{%
PI-explanations are related with abduction, and so are also referred
to as abductive explanations (AXp)~\cite{inms-aaai19}. More recently,
PI-explanations have been studied from a knowledge compilation
perspective~\cite{marquis-kr20}.}.
Formally, given $\mbf{v}=(v_1,\ldots,v_m)\in\mbb{F}$ with
$\kappa(\mbf{v})=c$, a PI-explanation (AXp) is any minimal subset
$\fml{X}\subseteq\fml{F}$ such that,
\begin{equation} \label{eq:axp}
  \forall(\mbf{x}\in\mbb{F}).
  \left[
    \bigwedge\nolimits_{i\in{\fml{X}}}(x_i=v_i)
    \right]
  \limply(\kappa(\mbf{x})=c)
\end{equation}
%
AXps can be viewed as answering a `Why?' question, i.e.\ why is some
prediction made given some point in feature space. A different view of
explanations is a contrastive explanation~\cite{miller-aij19}, which
answers a `Why Not?' question, i.e.\ which features can be changed to
change the prediction. A formal definition of contrastive explanation
is proposed in recent work~\cite{inams-aiia20}.
Given $\mbf{v}=(v_1,\ldots,v_m)\in\mbb{F}$ with $\kappa(\mbf{v})=c$, a
CXp is any minimal subset $\fml{Y}\subseteq\fml{F}$ such that,
\begin{equation} \label{eq:cxp}
  \exists(\mbf{x}\in\mbb{F}).\bigwedge\nolimits_{j\in\fml{F}\setminus\fml{Y}}(x_j=v_j)\land(\kappa(\mbf{x})\not=c) 
\end{equation}
Building on the results of R.~Reiter in model-based
diagnosis~\cite{reiter-aij87},~\cite{inams-aiia20} proves a minimal
hitting set (MHS) duality relation between AXps and CXps,
i.e.\ AXps are MHSes of CXps and vice-versa.

\autoref{sec:chkxp} studies the explanation membership problem, which
we define as follows:
\begin{definition}[AXp/CXp Membership Problem]
  Given $\mbf{v}\in\mbb{F}$ and $i\in\fml{F}$, with
  $\kappa(\mbf{v})=c\in\fml{K}$, the AXp (resp.~CXp) membership
  problem is to decide whether there exists an AXp (resp.~CXp)
  $\fml{Z}\subseteq\fml{F}$ with $i\in\fml{Z}$.
\end{definition}
One can understand the importance of deciding explanation membership
in settings where the number of explanations is very large, and we
seek to understand whether some feature (given its assigned value) can
be relevant for some prediction.
Duality between explanations~\cite{inams-aiia20} yields the following
result.

\begin{proposition} \label{prop:dualxp}
  Given $\mbf{v}\in\mbb{F}$, there exists an AXp
  $\fml{X}\subseteq\fml{F}$ with $i\in\fml{X}$ iff there exists a
  CXp $\fml{Y}\subseteq\fml{F}$ with $i\in\fml{Y}$.
\end{proposition}

\paragraph{Decision trees, diagrams and graphs.}

A decision tree $\fml{T}$ is a directed acyclic graph having at most
one path between every pair of nodes. $\fml{T}$ has a root node,
characterized by having no incoming edges. All other nodes have one
incoming edge. We consider univariate decision trees (as opposed to
multivariate decision trees~\cite{utgoff-ml95}), each non-terminal
node is associated with a single feature $x_i$.
Each edge is labeled with a literal, relating a feature (associated
with the edge's starting node) with some values (or range of values)
from the feature's domain. We will consider literals to be of the form
$x_i\Join{\mbb{E}_i}$, where $\Join\:\in\{\in\}$. $x_i$ is a variable
that denotes the value taken by feature $i$, whereas
$\mbb{E}_i\subseteq\mbb{D}_i$ is a subset of the domain of feature $i$.
The type of literals used to label the edges of a DT allows the
representation of the DTs generated by a wide range of decision
tree learners (e.g.~\cite{utgoff-ml97}).
(The syntax of the literals could be enriched. For example,
we could use $\Join\:\in\{\in,\not\in\}$, or for categorical features
we could use $\Join\:\in\{=,\not=\}$, in which case we would need to
allow for multi-tree variants of DTs~\cite{zeger-tit11}. However,
these alternatives would not change the results in the paper.
As formalized later, the literals associated with the outgoing edges
are assumed to be mutually inconsistent.
Finally, each terminal node is associated with a value from
$\fml{K}$.

Throughout the paper, we will use the following example of a DT as the
first running example.

\begin{figure}[t]
  \begin{center}
    \scalebox{0.875}{\tikzset{every label/.style={xshift=-0.15ex,
  yshift=-6.15ex,
  text width=1ex,
  align=right, inner sep=1pt, font=\tiny, text=midblue}}
\tikzset{tlabel/.style={xshift=0.25ex, yshift=2ex, text width=1ex,
    align=right, inner sep=1pt, font=\tiny, text=midblue}}
\tikzset{EV/.style = {font=\scriptsize}}
\forestset{
  BDT/.style={
    for tree={
      l=1.5cm,s sep=1.25cm,
      if n children=0{}{circle},
      draw,
      edge={
        my edge
      },
    }
  },
}
\begin{forest}
  BDT
  [$x_{3}$, label={1} 
    [$x_{1}$, label={2}, 
      edge label={node[EV,near start,left,xshift=-2pt] {$\in\{\tn{N}\}$}}
      [{\footnotesize\color{midgreen}\textbf{T}},
        label={[xshift=-0.05ex,yshift=1.725ex]4},
        edge label={node[EV,near start,left,xshift=-0pt] {$\in\{\tn{O}\}$}}
      ]
      [$x_{2}$, label={5}, 
        edge label={node[EV,near start,right,xshift=0pt] {$\in\{\tn{W},\tn{T}\}$}}
        [$x_{1}$, label={8}, 
          edge label={node[EV,near start,left,xshift=-0pt] {$\in\{\tn{H}\}$}}
          [{\footnotesize\color{midblue}\textbf{L}},
            label={[xshift=-0.05ex,yshift=1.725ex]12},
            edge label={node[EV,near start,left,xshift=-0pt] {$\in\{\tn{T}\}$}}
          ]
          [{\footnotesize\color{midred}\textbf{N}}, 
            label={[xshift=-0.05ex,yshift=1.725ex]13},
            edge label={node[EV,near start,right,xshift=0pt] {$\in\{\tn{W}\}$}}
          ]
        ]
        [{\footnotesize\color{midred}\textbf{N}},
          label={[xshift=-0.05ex,yshift=1.725ex]9},
          edge label={node[EV,near start,right,xshift=0pt] {$\in\{\tn{L},\tn{M}\}$}}
        ]
      ]
    ]
    [$x_4$, label={3}, edge=very thick,
      edge label={node[EV,near start,right,xshift=2pt] {$\in\{\tn{Y}\}$}}
      [{\footnotesize\color{midblue}\textbf{L}},
        label={[xshift=-0.05ex,yshift=1.725ex]6},
        edge label={node[EV,near start,left,xshift=-0pt] {$\in\{\tn{E}\}$}}
      ]
      [$x_{1}$, label={7}, edge=very thick,
        edge label={node[EV,near start,right,xshift=0pt] {$\in\{\tn{P},\tn{F}\}$}}
        [{\footnotesize\color{midgreen}\textbf{T}},
          label={[xshift=-0.05ex,yshift=1.725ex]10}, edge=very thick,
          edge label={node[EV,near start,left,xshift=-0pt] {$\in\{\tn{W},\tn{O}\}$}}
        ]
        [$x_{2}$, label={11}, 
          edge label={node[EV,near start,right,xshift=0pt] {$\in\{\tn{T}\}$}}
          [{\footnotesize\color{midgreen}\textbf{T}},
            label={[xshift=-0.05ex,yshift=1.725ex]14},
            edge label={node[EV,near start,left,xshift=-0pt] {$\in\{\tn{H}\}$}}
          ]
          [{\footnotesize\color{midblue}\textbf{L}}, 
            label={[xshift=-0.05ex,yshift=1.725ex]15},
            edge label={node[EV,near start,right,xshift=0pt] {$\in\{\tn{L},\tn{M}\}$}}
          ]
        ]
      ]
    ]
  ]
\end{forest}}
  \end{center}
  \caption{Example DT, $\mbf{v}=(\tn{O},\tn{L},\tn{Y},\tn{P})$ and
    $\kappa(\mbf{v})=\tn{\textcolor{midgreen}{\textbf{T}}}$} \label{fig:dt01a}
\end{figure}

\begin{example} \label{ex:dt01a}
  For the DT in~\autoref{fig:dt01a}, $\fml{F}=\{1,2,3,4\}$, denoting
  respectively Age ($\in\{\tn{W},\tn{T},\tn{O}\})$, Income
  ($\in\{\tn{L},\tn{M},\tn{H}\}$), Student ($\in\{\tn{N},\tn{Y}\}$)
  and Credit Rating ($\in\{\tn{P},\tn{F},\tn{E}\}$).
  The prediction is the type of hardware bought, with $\tn{N}$
  denoting No Hardware, $\tn{T}$ denoting a Tablet and $\tn{L}$
  denoting a Laptop.
  For Age, $\tn{W}$, $\tn{T}$ and $\tn{O}$ denote, respectively,
  $\tn{Age}<30$ (tWenties or younger), $30\le\tn{Age}<40$ (Thirties)
  and $40\le\tn{Age}$ (forties or Older).
  For Income, $\tn{L}$, $\tn{M}$, $\tn{H}$ denote, respectively,
  (L)ow, (M)edium, and (H)igh.
  For Student, $\tn{N}$ denotes not a student and $\tn{Y}$ denotes a
  student.
  Finally, for Credit Rating, $\tn{P}$, $\tn{F}$ and $\tn{E}$ denote,
  respectively, (P)oor, (F)air and (E)xcellent.
  For the instance $\mbf{v}=(\tn{O},\tn{L},\tn{Y},\tn{P})$,
  with prediction $\tn{T}$ (i.e.\ Tablet), 
  the consistent path is shown highlighted. 
\end{example}

The paper also considers reduced ordered  binary decision diagrams
(OBDDs)~\cite{bryant-tcomp86,wegener-bk00,darwiche-jair02}, as well as
their multi-valued
variant, i.e.\ reduced ordered multi-valued decision diagrams
(OMDDs)~\cite{brayton-iccad90,brayton-tr90,hooker-bk16}. (We will also
briefly mention connections with deterministic branching programs
(DBPs)~\cite{wegener-bk00}.)
For OBDDs, features must 
be boolean, and so 
each edge is labeled with either 0 or 1 (we could
instead use $\in\{0\}$ and $\in\{1\}$, respectively). In contrast with
DTs, features in OBDDs must also be ordered.
For OMDDs, features takes discrete values, and are also ordered. In
this case, we label edges the same way we label edges in DTs,
i.e.\ using set membership (and non-membership). (For simplicity, if
all edges have a single option, we just label the edge with the
value.)
Moreover, we will use the following example of an OMDD as the paper's
second running example.

\begin{figure}[t]
  \begin{center}
    \scalebox{0.875}{
\tikzset{>=Stealth}
\tikzset{tlabel/.style={xshift=-0.175ex, yshift=0.25ex, text width=1.0ex,
    align=right, inner sep=1pt, font=\tiny, text=midblue}}
\tikzset{blabel/.style={xshift=-0.1575ex, yshift=-0.1375ex, text width=1.25ex,
    align=right, inner sep=1pt, font=\tiny, text=midblue}}
\tikzset{rlabel/.style={xshift=0.25ex, yshift=0ex, text width=1ex,
    align=right, inner sep=1pt, font=\tiny, text=midblue,align=left}}
\tikzset{llabel/.style={xshift=-3.25ex, yshift=2.775ex, text width=1.25ex,
    align=right, inner sep=1pt, font=\tiny, text=midblue,align=right}}
\tikzset{rblabel/.style={xshift=-0.175ex, yshift=-1.35ex, text width=1ex,
    align=right, inner sep=1pt, font=\tiny, text=midblue,align=left}}
\begin{tikzpicture}[->,%
    EV/.style = {font=\footnotesize},
    node distance={2.0cm}, thin, main/.style = {draw, circle}] 
  \node[main] (1) {$x_3$};
  \node[main] (2) [below left of=1]  {$x_2$};
  \node[main] (3) [below right of=1] {$x_2$};
  \node[main] (4) [below of=2]       {$x_1$};
  \node[main] (5) [below of=3]       {$x_1$};
  \node[main] (7) [below of=4]       {$\color{midgreen}{\tn{\textbf{G}}}$};
  \node[main] (6) [left of=7]        {$\color{midred}{\tn{\textbf{R}}}$};
  \node[main] (8) [below of=5]       {$\color{midblue}{\tn{\textbf{B}}}$};
  \node[above = 0 of 1,style=tlabel]  (l1) {1};
  \node[right = 0 of 2,style=rblabel] (l2) {2};
  \node[right = 0 of 3,style=rblabel] (l3) {3};
  \node[right = 0 of 4,style=rblabel] (l4) {4};
  \node[right = 0 of 5,style=rblabel] (l5) {5};
  \node[right = 0 of 6,style=rblabel] (l6) {6};
  \node[right = 0 of 7,style=rblabel] (l7) {7};
  \node[right = 0 of 8,style=rblabel] (l8) {8};
  \draw[] (1) -- node [EV, near start, below=1pt]  {1} (2);
  \draw[] (1) edge[very thick] node [EV, near start, below=1pt]  {2} (3);
  \draw[] (1) edge[bend right=45] node[EV, near start, above=1pt]   {0} (6);
  \draw[] (2) edge[bend right=30] node [EV, near start, above=1pt]  {0} (6);
  \draw[] (2) -- node [EV, near start, left=0pt]  {1} (4);
  \draw[] (3) edge[bend right=30] node [EV, near start, below=0pt] {0} (6);
  \draw[] (3) edge[very thick] node [EV, near start, right=0pt] {1} (5);
  \draw[] (4) edge [bend left=10] node [EV, near start, above=1pt]  {0} (6);
  \draw[] (4) -- node [EV, near start, right=0pt] {1} (7);
  \draw[] (5) edge [bend left=8,very thick] node [EV, near start, above=1pt]   {0} (6);
  \draw[] (5) -- node [EV, near start, right=0pt] {1} (8);
\end{tikzpicture} }
  \end{center}
  \caption{Example OMDD, $\mbf{v}=(0,1,2)$ and
    $\kappa(\mbf{v})=\tn{\textcolor{midred}{\textbf{R}}}$} \label{fig:mdd01a}
\end{figure}

\begin{example} \label{ex:mdd01a}
  For the OMDD in~\autoref{fig:mdd01a}, $\fml{F}=\{1,2,3\}$,
  with $D_1=D_2=\{0,1\}$, $D_3=\{0,1,2\}$.
  The prediction is one of three classes
  $\fml{K}=\{\tn{R},\tn{G},\tn{B}\}$.
  For the instance $\mbf{v}=(0,1,2)$, 
  with prediction $\tn{R}$, 
  the consistent path is shown highlighted. 
\end{example}

Over the years, different works proposed the use of some sort of
decision diagrams as an alternative to decision
trees,
e.g.~\cite{mercer-tassp89,oliver-tr92,oliveira-ml96,baesens-esa04,ccimspp-date21}. 
This paper considers \emph{decision graphs}~\cite{oliver-tr92}, which
can be viewed as a generalization of DTs, OBDDs, OMDDs, etc. 
\begin{definition}[Decision Graph (DG)]
  A DG $\fml{G}$ is a 4-tuple $\fml{G}=(G,\varsigma,\phi,\lambda)$
  where,
  \begin{enumerate}[nosep]
    \item $G=(V,E)$ is a Directed Acyclic Graph (DAG)  with a single
      source (or root) node.
    \item $V$ is partitioned into terminal ($T$) and non-terminal
      ($N$) nodes.
      For every $p\in{N}$, $\outdeg(p)>0$.
      For every $q\in{T}$, $\outdeg(q)=0$.
      ($\outdeg$ denotes the outdegree of a node).
    \item
      $\varsigma:T\to\fml{K}$ maps each terminal node into a class.
    \item
      $\phi:N\to\fml{F}$ maps each non-terminal node into a feature.
    \item $\lambda:E\to\mfrak{L}$, where $\mfrak{L}$ denotes the set
      of all literals of the form $x_i\in{\mbb{E}_i}$ for
      $\mbb{E}_i\subseteq\mbb{D}_i$, where $i$ is the feature
      associated with the edge's starting node.
  \end{enumerate}
  Furthermore, the following assumptions are made with respect to
  DGs\footnote{%
    The importance of these assumptions must be highlighted.
    Whereas for OBDDs/OMDDs these assumptions are guaranteed by
    construction, this is not the case with DTs nor in general with
    DGs.
    In the literature, one can find examples of decision trees with
    inconsistent paths~(e.g.\ \cite[Fig.~4]{valdes-sr16}) but also
    decision trees exhibiting dead-ends, i.e.\ DTs for which the
    classification function is not
    total~(e.g.\ \cite[Fig.~8.1]{duda-bk01}).} (where for node
  $r\in{N}$, with $\phi(r)=i$, $\mbb{C}_i$ denotes the set of values
  of feature $i$ which are consistent with any path connecting the
  root to $r$):
  \begin{enumerate}[nosep,label=\roman*.]
  \item The literals associated with the outgoing edges of each node
    $r\in{N}$ represent a partition of $\mbb{C}_i$.
  \item Every path $R_k$ of DG, that connects the root node to a
    terminal node, is not inconsistent.
  \end{enumerate}
\end{definition}

It is straightforward to conclude that any classifier defined on DTs,
OBDDs or OMDDs can be represented as a DG. (The same claim can also
be made for DBPs.) Also, whereas OBDDs and OMDDs are read-once
(i.e.\ each feature is tested at most once along a path), DTs (and
general DGs) need not be read-once. Hence, DGs impose no restriction
on the number of times a feature is tested along a path, as long as
the literals are consistent.
Moreover, the definition of DG (and the associated assumptions)
ensures that,

\begin{proposition}
  For any $\mbf{v}\in\mbb{F}$ there exists exactly one terminal node
  which is connected to the root node of the DG by path(s) such that
  each edge in such path(s) is consistent with the feature values
  given by $\mbf{v}$.
\end{proposition}

\begin{proposition}
  For any terminal node $q$ of a DG, there exists at least one point
  $\mbf{v}\in\mbb{F}$ such there is a consistent path from the root to
  $q$.
\end{proposition}

\section{Explanation Graphs} \label{sec:xpg}

A difficulty with reasoning about explanations for DTs, DGs or OBDDs,
but also for their multi-valued variants (and also in the case of
other examples of ML models), is the multitude of cases that one needs
to consider. For the concrete case of OBDDs, features are restricted
to be boolean. However, for DTs and DGs, features can be boolean,
categorical, integer or real. Moreover, for OMDDs, features can be
boolean, categorical or integer.
Also, it is often the case that $|\fml{K}|>2$.
Explanation graphs are a graph representation that abstracts
away all the details that are effectively unnecessary for computing
AXps or CXps. In turn, this facilitates the construction of unified
explanation procedures.

\begin{definition}[Explanation Graph (XpG)]
  An 
  XpG is a 5-tuple
  $\fml{D}=(G_{\fml{D}},S,\upsilon,\alpha_{V},\alpha_{E})$, where:
  \begin{enumerate}[nosep]
  \item $G_{\fml{D}}=(V_{\fml{D}},E_{\fml{D}})$ is a labeled DAG, such
    that:
    \begin{itemize}[nosep]
    \item $V_{\fml{D}}=T_{\fml{D}}\cup{N_{\fml{D}}}$ is the set of
      nodes, partitioned into the terminal nodes $T_{\fml{D}}$
      (with $\outdeg(q)=0$, $q\in{T_{\fml{D}}}$) and the non-terminal
      nodes $N_{\fml{D}}$ (with $\outdeg(p)>0$, $p\in{N_{\fml{D}}}$);
    \item $E_{\fml{D}}\subseteq{V_{\fml{D}}}\times{V_{\fml{D}}}$ is
      the set of (directed) edges.
    \item $G_{\fml{D}}$ is such that there is a single node with
      indegree equal to 0, i.e.\ the root (or source) node.
    \end{itemize}
  \item $S=\{s_1,\ldots,s_m\}$ is a set of variables;
  \item $\upsilon:N_{{\fml{D}}}\to{S}$ is a total function mapping
    each non-terminal node to one variable in $S$.
  \item $\alpha_V:V_{\fml{D}}\to\{0,1\}$ labels nodes with one of two
    values.\\
    ($\alpha_{V}$ is required to be defined only for terminal nodes.)
  \item $\alpha_E:E_{\fml{D}}\to\{0,1\}$ labels edges with one of two
    values.
  \end{enumerate}
  In addition, an XpG $\fml{D}$ must respect the following properties:
  \begin{enumerate}[nosep,label=\roman*.]
  \item For each non-terminal node, there is at most one outgoing edge
    labeled 1; all other outgoing edges are labeled 0.
  \item There is exactly one terminal node $t\in{T}$ labeled 1 that
    can be reached from the root node with (at least) one path of
    edges labeled 1.
  \end{enumerate}
\end{definition}
We refer to a \emph{tree XpG} when the DAG associated with the XpG is
a tree.
Given a DG $\fml{G}$ and an instance $(\mbf{v},c)$, the (unique)
mapping to an XpG is obtained as follows:
\begin{enumerate}[nosep]
\item The same DAG is used.
\item Terminal nodes labeled $c$ in $\fml{G}$ are labeled 1 in
  $\fml{D}$. Terminal nodes labeled $c'\not=c$ in $\fml{G}$ are
  labeled 0 in $\fml{D}$.
\item A non-terminal node associated with feature $i$ in $\fml{G}$
  is associated with $s_i$ in $\fml{D}$.
\item Any edge labeled with a literal that is consistent with
  $\mbf{v}$ in $\fml{G}$ is labeled 1 in $\fml{D}$. Any edge labeled
  with a literal that is not consistent with $\mbf{v}$ in $\fml{G}$ is
  labeled 0 in $\fml{D}$.
\end{enumerate}
Since we can represent DTs, OBDDs or OMDDs with DGs, then the
construction above ensures that we can also create XpG's for any of
these classifiers.

The following examples illustrate the construction of XpG's for the
paper's two running examples.

\begin{example} \label{ex:dt01b}
  For the DT of \autoref{ex:dt01a} (shown in~\autoref{fig:dt01a},
  given the instance $(\mbf{v}=(\tn{O},\tn{L},\tn{Y},\tn{P}), \tn{T})$,
  and letting $S=(s_1,s_2,s_3,s_4)$, with each $s_i$ associated with
  feature $i$, the resulting XpG is shown in \autoref{fig:dt01b}.
\end{example}

\begin{figure}[t]
  \begin{center}
    \scalebox{0.875}{\tikzset{every label/.style={xshift=-0.35ex,
  yshift=-6ex,
  text width=1ex,
  align=right, inner sep=1pt, font=\tiny, text=midblue}}
\tikzset{tlabel/.style={xshift=0.25ex, yshift=2ex, text width=1ex,
    align=right, inner sep=1pt, font=\tiny, text=midblue}}
\tikzset{EVR/.style = {font=\scriptsize\bfseries,text=midred}}
\tikzset{EVG/.style = {font=\scriptsize\bfseries,text=midgreen}}
\forestset{
  BDT/.style={
    for tree={
      l=1.5cm,s sep=1.25cm,
      if n children=0{}{circle},
      draw,
      edge={
        my edge
      },
    }
  },
}
\begin{forest}
  BDT
  [$s_{3}$, label={1} 
    [$s_{1}$, label={2}, 
      edge label={node[EVR,near start,left,xshift=-2pt] {0}}
      [{\footnotesize\color{midgreen}\textbf{1}},
        label={[xshift=-0.05ex,yshift=1.725ex]4},
        edge label={node[EVG,near start,left,xshift=-0.5pt] {1}}
      ]
      [$s_{2}$, label={5},
        edge label={node[EVR,near start,right,xshift=0.5pt] {0}}
        [$s_{1}$, label={8}, 
          edge label={node[EVR,near start,left,xshift=-0.5pt] {0}}
          [{\footnotesize\color{midred}\textbf{0}},
            label={[xshift=-0.05ex,yshift=1.725ex]12},
            edge label={node[EVR,near start,left,xshift=-0.5pt] {0}}
          ]
          [{\footnotesize\color{midred}\textbf{0}}, 
            label={[xshift=-0.05ex,yshift=1.725ex]13},
            edge label={node[EVR,near start,right,xshift=0.5pt] {0}}
          ]
        ]
        [{\footnotesize\color{midred}\textbf{0}},
          label={[xshift=-0.05ex,yshift=1.725ex]9},
          edge label={node[EVG,near start,right,xshift=0.5pt] {1}}
        ]
      ]
    ]
    [$s_4$, label={3},
      edge label={node[EVG,near start,right,xshift=2pt] {1}}
      [{\footnotesize\color{midred}\textbf{0}},
        label={[xshift=-0.05ex,yshift=1.725ex]6},
        edge label={node[EVR,near start,left,xshift=-0.5pt] {0}}
      ]
      [$s_{1}$, label={7}, 
        edge label={node[EVG,near start,right,xshift=0.5pt] {1}}
        [{\footnotesize\color{midgreen}\textbf{1}},
          label={[xshift=-0.05ex,yshift=1.725ex]10},
          edge label={node[EVG,near start,left,xshift=-0.5pt] {1}}
        ]
        [$s_{2}$, label={11}, 
          edge label={node[EVR,near start,right,xshift=0.5pt] {0}}
          [{\footnotesize\color{midgreen}\textbf{1}},
            label={[xshift=-0.05ex,yshift=1.725ex]14},
            edge label={node[EVR,near start,left,xshift=-0.5pt] {0}}
          ]
          [{\footnotesize\color{midred}\textbf{0}}, 
            label={[xshift=-0.05ex,yshift=1.725ex]15},
            edge label={node[EVG,near start,right,xshift=0.5pt] {1}}
          ]
        ]
      ]
    ]
  ]
\end{forest}}
  \end{center}
  \caption{XpG for the DT in~\autoref{fig:dt01a}, given
    $\mbf{v}=(\tn{O},\tn{L},\tn{Y},\tn{P})$} \label{fig:dt01b}
\end{figure}

\begin{example} \label{ex:mdd01b}
  For the OMDD of ~\autoref{ex:mdd01a} (shown
  in~\autoref{fig:mdd01a}), given the instance $((0,1,2), \tn{R})$,
  and letting $S=(s_1,s_2,s_3)$, with each $s_i$ associated with
  feature $i$, the resulting XpG is shown in \autoref{fig:mdd01b}.
\end{example}

\begin{figure}[t]
  \begin{center}
    \scalebox{0.875}{
\tikzset{>=Stealth}
\tikzset{tlabel/.style={xshift=-0.175ex, yshift=0.25ex, text width=1.0ex,
    align=right, inner sep=1pt, font=\tiny, text=midblue}}
\tikzset{blabel/.style={xshift=-0.1575ex, yshift=-0.1375ex, text width=1.25ex,
    align=right, inner sep=1pt, font=\tiny, text=midblue}}
\tikzset{rlabel/.style={xshift=0.25ex, yshift=0ex, text width=1ex,
    align=right, inner sep=1pt, font=\tiny, text=midblue,align=left}}
\tikzset{llabel/.style={xshift=-3.25ex, yshift=2.775ex, text width=1.25ex,
    align=right, inner sep=1pt, font=\tiny, text=midblue,align=right}}
\tikzset{rblabel/.style={xshift=-0.175ex, yshift=-1.35ex, text width=1ex,
    align=right, inner sep=1pt, font=\tiny, text=midblue,align=left}}
\begin{tikzpicture}[->,%
    EV/.style = {font=\footnotesize\bfseries,},
    node distance={2.0cm}, thin, main/.style = {draw, circle}] 
  \node[main] (1) {$s_3$};
  \node[main] (2) [below left of=1]  {$s_2$};
  \node[main] (3) [below right of=1] {$s_2$};
  \node[main] (4) [below of=2]       {$s_1$};
  \node[main] (5) [below of=3]       {$s_1$};
  \node[main] (7) [below of=4]       {$\color{midred}{\tn{\textbf{0}}}$};
  \node[main] (6) [left of=7]        {$\color{midgreen}{\tn{\textbf{1}}}$};
  \node[main] (8) [below of=5]       {$\color{midred}{\tn{\textbf{0}}}$};
  \node[above = 0 of 1,style=tlabel]  (l1) {1};
  \node[right = 0 of 2,style=rblabel] (l2) {2};
  \node[right = 0 of 3,style=rblabel] (l3) {3};
  \node[right = 0 of 4,style=rblabel] (l4) {4};
  \node[right = 0 of 5,style=rblabel] (l5) {5};
  \node[right = 0 of 6,style=rblabel] (l6) {6};
  \node[right = 0 of 7,style=rblabel] (l7) {7};
  \node[right = 0 of 8,style=rblabel] (l8) {8};
  \draw[] (1) -- node [EV, near start, below=1pt]  {\color{midred}{0}} (2);
  \draw[] (1) -- node [EV, near start, below=1pt]  {\color{midgreen}{1}} (3);
  \draw[] (1) edge[bend right=45] node[EV, near start, above=1pt]   {\color{midred}{0}} (6);
  \draw[] (2) edge[bend right=30] node [EV, near start, above=1pt]  {\color{midred}{0}} (6);
  \draw[] (2) -- node [EV, near start, left=0pt]  {\color{midgreen}{1}} (4);
  \draw[] (3) edge [bend right=30] node [EV, near start, below=0pt] {\color{midred}{0}} (6);
  \draw[] (3) -- node [EV, near start, right=0pt] {\color{midgreen}{1}} (5);
  \draw[] (4) edge [bend left=10] node [EV, near start, above=1pt]  {\color{midgreen}{1}} (6);
  \draw[] (4) -- node [EV, near start, right=0pt] {\color{midred}{0}} (7);
  \draw[] (5) edge [bend left=8] node [EV, near start, above=1pt]   {\color{midgreen}{1}} (6);
  \draw[] (5) -- node [EV, near start, right=0pt] {\color{midred}{0}} (8);
\end{tikzpicture} }
  \end{center}
  \caption{XpG for the OMDD of~\autoref{fig:mdd01a}, given
    $\mbf{v}=(0,1,2)$} \label{fig:mdd01b}
\end{figure}

\paragraph{Evaluation of XpG's.}

Given an XpG $\fml{D}$, let $\mbb{S}=\{0,1\}^m$, i.e.\ the set of
possible assignments to the variables in $S$. The evaluation function
of the XpG, $\sigma_{\fml{D}}:\mbb{S}\to\{0,1\}$, is based on the
auxiliary \emph{activation} function
$\varepsilon:\mbb{S}\times{V_{\fml{D}}}\to\{0,1\}$.
Moreover, for a point $\mbf{s}\in\mbb{S}$, $\sigma_{\fml{D}}$ and
$\varepsilon$ are defined as follows:
\begin{enumerate}[nosep]
\item If $r$ is the root node of $G_{\fml{D}}$, then
  $\varepsilon(\mbf{s},r)=1$.
\item Let $p\in\parent(r)$ (i.e.\ a node can have multiple parents)
  and let $s_i=\upsilon(p)$.
  $\varepsilon(\mbf{s},r)=1$ iff $\varepsilon(\mbf{s},p)=1$ and either
  $\alpha_{E}(p,r)=1$ or $s_i=0$, i.e.
  \begin{equation}\small
  \varepsilon(\mbf{s},r)\equiv
  \bigvee_{\substack{p\in\parent(r)\\\land\neg\alpha_{E}(p,r)}}\left(\varepsilon(\mbf{s},p)\land\neg{s_i}\right)
  \bigvee_{\substack{p\in\parent(r)\\\land\alpha_{E}(p,r)}}\varepsilon(\mbf{s},p)
  \end{equation}
\item $\sigma_{\fml{D}}(\mbf{s})=1$ iff for every terminal node
  $t\in{T_{\fml{D}}}$, with $\alpha_{V}(t)=0$, it is also the case that
  $\varepsilon(\mbf{s},t)=0$, i.e.
  \begin{equation}\small
    \sigma_{\fml{D}}(\mbf{s})\equiv
    \bigwedge\nolimits_{t\in{T_{\fml{D}}}\land\neg\alpha_{V}(t)}\neg\varepsilon(\mbf{s},t)
  \end{equation}
\end{enumerate}
Observe that terminal nodes labeled 1 are irrelevant for defining the
evaluation function. Their existence is implicit (i.e.\ at least one
terminal node with label 1 must exist and be reachable from the root
when all the $s_i$ variables take value 1), but the evaluation of
$\sigma_{\fml{D}}$ is oblivious to their existence.
Furthermore, and as noted above, we must have
$\sigma_{\fml{D}}(1,\ldots,1)=1$.
If the graph has some terminal node labeled 0, then
$\sigma_{\fml{D}}(0,\ldots,0)=0$.

\begin{example} \label{ex:dt01c}
  For the DT of~\autoref{fig:dt01a}, and given the XpG
  of~\autoref{fig:dt01b}, the evaluation function is defined as
  follows:
  \begin{small}\[
  \sigma_{\fml{D}}(\mbf{s})\lequiv%
  \left(\bigwedge\nolimits_{r\in\{6,9,12,13,15\}}\neg\varepsilon(\mbf{s},r)\right)
  \]\end{small}
  with,
  \begin{small}\[
  \begin{array}{l}
    \left[\varepsilon(\mbf{s},1)\lequiv1\right]
    \,\land\,
    \left[\varepsilon(\mbf{s},2)\lequiv\varepsilon(\mbf{s},1)\land\neg{s_3}\right]
    \,\land\,
    \\
    \left[\varepsilon(\mbf{s},3)\lequiv\varepsilon(\mbf{s},1)\right]
    \,\land\,
    \left[\varepsilon(\mbf{s},5)\lequiv\varepsilon(\mbf{s},2)\land\neg{s_1}\right]
    \,\land\,
    \\
    \left[\varepsilon(\mbf{s},6)\lequiv\varepsilon(\mbf{s},3)\land\neg{s_4}\right]
    \,\land\,
    \left[\varepsilon(\mbf{s},7)\lequiv\varepsilon(\mbf{s},3)\right]
    \,\land\,
    \\
    \left[\varepsilon(\mbf{s},8)\lequiv\varepsilon(\mbf{s},5)\land\neg{s_2}\right]
    \,\land\,
    \left[\varepsilon(\mbf{s},9)\lequiv\varepsilon(\mbf{s},5)\right]
    \,\land\,
    \\
    \left[\varepsilon(\mbf{s},11)\lequiv\varepsilon(\mbf{s},7)\land\neg{s_1}\right]
    \,\land\,
    \left[\varepsilon(\mbf{s},12)\lequiv\varepsilon(\mbf{s},8)\land\neg{s_1}\right]
    \,\land\,
    \\
    \left[\varepsilon(\mbf{s},13)\lequiv\varepsilon(\mbf{s},8)\land\neg{s_1}\right]
    \,\land\,
    \left[\varepsilon(\mbf{s},15)\lequiv\varepsilon(\mbf{s},11)\right]
    \\
  \end{array}
  \]\end{small}
  (where, for simplicity and for reducing the number of parenthesis,
  the operator $\land$ has precedence over the operator $\lequiv$.)
  Observe that $\sigma_{\fml{D}}(1,1,1,1)=1$
  and $\sigma_{\fml{D}}(0,0,0,0)=0$.
\end{example}

\begin{example}  \label{ex:mdd01c}
  For the OMDD of~\autoref{fig:mdd01a}, and given the XpG
  of~\autoref{fig:mdd01b}, the evaluation function is defined as
  follows:
  \begin{small}\[
  \sigma_{\fml{D}}(\mbf{s})\lequiv%
  \left(\bigwedge\nolimits_{r\in\{7,8\}}\neg\varepsilon(\mbf{s},r)\right)
  \]\end{small}
  with,
  \begin{small}\[
  \begin{array}{l}
    \left[\varepsilon(\mbf{s},1)\lequiv1\right]
    \,\land\,
    \left[\varepsilon(\mbf{s},2)\lequiv\varepsilon(\mbf{s},1)\land\neg{s_3}\right]
    \,\land\,
    \\
    \left[\varepsilon(\mbf{s},3)\lequiv\varepsilon(\mbf{s},1)\right]
    \,\land\,
    \left[\varepsilon(\mbf{s},4)\lequiv\varepsilon(\mbf{s},2)\right]
    \,\land\,
    \\
    \left[\varepsilon(\mbf{s},5)\lequiv\varepsilon(\mbf{s},3)\right]
    \,\land\,
    \left[\varepsilon(\mbf{s},7)\lequiv\varepsilon(\mbf{s},4)\land\neg{s_1}\right]
    \,\land\,
    \\
    \left[\varepsilon(\mbf{s},8)\lequiv\varepsilon(\mbf{s},5)\land\neg{s_1}\right]
    \\
  \end{array}
  \]\end{small}
  Again, we have $\sigma_{\fml{D}}(1,1,1,1)=1$ and $\sigma_{\fml{D}}(0,0,0,0)=0$.
\end{example}

\paragraph{Properties of XpG's.}
The definition of $\sigma_{\fml{D}}$ is such that the evaluation
function is monotone (where we define $0\preceq1$,
$\mbf{s}_1\preceq\mbf{s}_2$ if for all $i$, $s_{1,i}\preceq{s_{2,i}}$,
and for monotonicity we require
$\mbf{s}_1\preceq\mbf{s}_2\limply\sigma_{\fml{D}}(\mbf{s}_1)\preceq\sigma_{\fml{D}}(\mbf{s}_2)$.

\begin{proposition}
  Given an XpG $\fml{D}$, $\sigma_{\fml{D}}$
  is monotone.
\end{proposition}

\begin{Proof}[Sketch]
  Observe that $\varepsilon$ is monotone (and negative) on
  $\mbf{s}\in\mbb{S}$, and $\sigma_{\fml{D}}$ is monotone (and
  negative) on $\varepsilon$.
  Hence, $\sigma_{\fml{D}}$ is monotone (and positive) on $\mbf{s}$.
\end{Proof}

Given the definition of $\sigma_{\fml{D}}$, any PI will consist of a
conjunction of positive literals~\cite{crama-bk11}.
Furthermore, we can view an XpG as a classifier, mapping features
$\{1,\ldots,m\}$ (each feature $i$ associated with a variable
$s_i\in{S}$) into $\{0,1\}$, with instance $((1,\ldots,1),1)$.
As a result, we can compute the AXps and CXps of an XpG $\fml{D}$
(given the instance $((1,\ldots,1),1)$).

\begin{example}
  Observe that by setting $s_2=s_3=0$, we still guarantee that
  $\sigma_{\fml{D}}(1,0,0,1)=1$. However, setting either $s_1=0$ or
  $s_4=0$, will cause $\sigma_{\fml{D}}$ to change value.
  Hence, one AXp for the XpG is $\{1,4\}$. With respect to the original
  instance $((\tn{O},\tn{L},\tn{Y},\tn{P}),\tn{T})$, selecting
  $\{1,4\}$ indicates that $(x_1=\tn{O})\land(x_4=\tn{P})$ (i.e.\ Age
  in the forties or Older and a Credit Rating of Poor) suffices for
  the prediction of $\tn{T}$.
\end{example}

\begin{example}
  With respect to \autoref{ex:mdd01c}, we can observe that $s_2$ is
  not used for defining $\sigma_{\fml{D}}$. Hence, it can be set to
  0. Also, as long as $s_1=1$, the prediction will remain unchanged.
  Thus, we can also set $s_3$ to 0. As a result, one AXp is $\{1\}$.
  With respect to the original instance
  $((x_1,x_2,x_3),c)=((0,1,2),\tn{R})$, selecting $\{1\}$ indicates that
  $x_1=0$ suffices for the prediction of \tn{R}.
\end{example}

As suggested by the previous discussion and examples, we have the
following result.

\begin{proposition}
  There is a one-to-one mapping between AXps and CXps of
  $\sigma_{\fml{D}}$ and the AXps and CXps of the original
  classification problem (and instance) from which the XpG $\fml{D}$
  is obtained.
\end{proposition}

\begin{Proof}[Sketch]
  The construction of the XpG from a DG ensures that for any node in
  the XpG, if $\varepsilon(\mbf{s},r)=1$, then there exists some
  assignment to the features corresponding to unset variables, such
  that there is one consistent path in the DG from the root to $r$.
  Thus, if for some pick of unset variables, we have that
  $\varepsilon(\mbf{s},q)=1$, for some $q\in{T_{\fml{D}}}$ with
  $\alpha_{V}(q)=0$, then that  guarantees that in the DG there is an
  assignment to the features associated with the unset variables, such
  that a prediction other than $c$ is obtained.
\end{Proof}

\section{Computing Explanations} \label{sec:xps}

It is well-known that prime implicants of monotone functions can be
computed in polynomial time
(e.g.~\cite{mundhenk-mfcs05,mundhenk-ic08}).
Moreover, whereas there are algorithms for finding one PI of a
monotone function in polynomial time, there is evidence that
enumeration of PIs cannot be achieved with polynomial
delay~\cite{khachiyan-dam99}.

Nevertheless, and given the fact that $\sigma_{\fml{D}}$ is defined on
a DAG, this paper proposes dedicated algorithms for computing one AXp
and one CXp which build on iterative graph traversals.
Furthermore, the MARCO algorithm~\cite{lpmms-cj16} is adapted to
exploit the algorithms for computing one AXp and one CXp, in the
process ensuring that AXps/CXps can be enumerated with exactly one
SAT oracle call per each computed explanation.

\subsection{Finding One XP} \label{ssec:onexp}

Different polynomial-time algorithms can be envisioned for finding one
prime implicant of an XpG (and also of a monotone function).
For the concrete case of $\sigma_{\fml{D}}$, we consider the
well-known deletion-based algorithm~\cite{chinneck-jc91}, which
iteratively removes literals from the implicant, and checks the value
of $\sigma_{\fml{D}}$ using the DAG representation.
(It is also plain that 
we could consider instead the algorithms
QuickXplain~\cite{junker-aaai04} or Progression~\cite{msjb-cav13}, or
any other algorithm for finding a minimal set subject to a monotone
predicate~\cite{msjm-aij17}.)

As highlighted in the running examples, if
$\sigma_{\fml{D}}(\mbf{u})=1$, for some $\mbf{u}\in\mbb{S}$, then in
the original classifier this means the prediction remains
unchanged. The only way we have to change the prediction is to allow
some features to take some other value from their domain. As a result,
we equate $s_i=1$ with declaring the original feature as \emph{set}
(or as fixed), whereas we equate $s_i=0$ with declaring the original
feature as \emph{unset} (or as free).
By changing some of the $S$ variables from 1 to 0, we are allowing
some of the features to take one value from their domains.
If we manage to change the value of the evaluation function to 0, this
means that in the original classifier there exists a pick of values to
the unset features which allows the prediction to change to some class
other than $c$.
As a result, the algorithms proposed in this section are solely based
on finding a subset maximal set of features declared free
(respectively, fixed), which is sufficient for the prediction not to
change (respectively, to change).

To enable the integration of the algorithms, the basic algorithms for
finding one XP are organized such that one XP is computed given a
starting seed.

\paragraph{Checking path to node with label 0.}
All algorithms are based on graph traversals, which check whether a
prediction of 0 can be reached given a set of value picks for the
variables in $S$.
This graph traversal algorithm is simple to envision, and is shown
in~\autoref{alg:cpath}.
\begin{algorithm}[t]
\begin{algorithmic}[1]
  \Input{XpG: $\fml{D}=(G_{\fml{D}},S,\upsilon,\alpha_{V},\alpha_{E})$;
    Ref set: ${R}\subseteq{S}$}
  \Procedure{$\cpath$}{$\fml{D},{R}$}
  \State{$\mbb{Q}\gets\initQ(\rootG(G_{\fml{D}}))$}
  \While{\NOT~$\emptyQ(\mbb{Q})$}
  \State{$(\mbb{Q},p)\gets\delQ(\mbb{Q})$}
  \If{$\isterm(G_{\fml{D}},p)$}
  \If{$\alpha_{V}(p)=0$}
  \State{\RETURN~{\True}}
  \EndIf
  \Else
  \State{$s_i\gets\upsilon(p)$}
  \ForAll{$q\in\childs(G_{\fml{D}},p)$}
  \If{$s_i\in{R}$~\OR~$\alpha_E(p,q)=1$}
  \State{$\mbb{Q}\gets\addQ(\mbb{Q},q)$}
  \EndIf
  \EndFor
  \EndIf
  \EndWhile
  \State{\RETURN~{\False}}
  \EndProcedure
\end{algorithmic}

  \caption{Check existence of path to 0-labeled terminal} 
  \label{alg:cpath}
\end{algorithm}
As can be observed, the algorithm returns 1 if a terminal labeled 0
can be reached. Otherwise, it returns 0. Variables in set $R$ serve to
ignore the values of outgoing edges of a node if the variable is in
$R$. The algorithm has a linear run time on the XpG's size.

\paragraph{Extraction of one AXp and one CXp given seed.}
\begin{algorithm}[t]
\begin{algorithmic}[1]
  \Input{XpG:
    $\fml{D}=(G_{\fml{D}},S,\upsilon,\alpha_{V},\alpha_{E})$;
    Seed set: ${A}\subseteq{S}$}
  \Procedure{$\axp$}{$\fml{D},{A}$}
  %
  \ForAll{\label{alg:axpe:ln01}$s_i\in{A}$}\Comment{Inv.:~$\NOT~\cpath(\fml{D},{S}{\setminus}{A})$}
  \If{\label{alg:axpe:ln02}\NOT~$\cpath(\fml{D},S\setminus({A}\setminus\{s_i\}))$}
  \State{\label{alg:axpe:ln03}${A}\gets{A}\setminus\{s_i\}$}
  \EndIf
  \EndFor
  \State{\label{alg:axpe:ln04}\RETURN~${A}$}
  \EndProcedure
\end{algorithmic}

  \caption{Extraction of one AXp given seed $A$} \label{alg:axpe} 
\end{algorithm}
Given a seed set $A\subseteq{S}$ of set variables, and so a set
$C={S}\setminus{A}$ of unset variables (which are \emph{guaranteed} to
be kept unset), \autoref{alg:axpe} drops variables from $A$
(i.e.\ makes variables unset, and so allows the original features to
take one of the values in their domains).
Since $\sigma_{\fml{D}}$ is monotone, the deletion-based algorithm is
guaranteed to find a subset-minimal set of fixed variables such that
the XpG evaluates to 1.

\begin{algorithm}[t]
\begin{algorithmic}[1]
  \Input{XpG:
    $\fml{D}=(G_{\fml{D}},S,\upsilon,\alpha_{V},\alpha_{E})$;
    Seed set: ${C}\subseteq{S}$}
  \Procedure{$\cxp$}{$\fml{D},{C}$}
  %
  \ForAll{\label{alg:cxpe:ln01}$s_i\in{C}$}\Comment{Inv.:~$\cpath(\fml{D},{C})$}
  \If{\label{alg:cxpe:ln02}$\cpath(\fml{D},{C}\setminus\{s_i\})$}
  \State{\label{alg:cxpe:ln03}${C}\gets{C}\setminus\{s_i\}$}
  \EndIf
  \EndFor
  \State{\label{alg:cxpe:ln04}\RETURN~{${C}$}}
  \EndProcedure
\end{algorithmic}

  \caption{Extraction of one CXp given seed $C$} \label{alg:cxpe} 
\end{algorithm}
Similarly, given a seed set $C\subseteq{S}$ of unset variables, and so
a set $A={S}\setminus{C}$ of set variables (which are
\emph{guaranteed} to be kept set), \autoref{alg:cxpe} drops variables
from $C$ (i.e.\ makes variables set, and so forces the original
features to take the value specified by the instance).

\subsection{Enumeration of Explanations} \label{ssec:allxp}

\begin{algorithm}[t]
  \begin{algorithmic}[1]
  \Input{XpG: $\fml{D}=(G_{\fml{D}},S,\upsilon,\alpha_{V},\alpha_{E})$}
  \Procedure{$\mathsf{Enumerate}$}{$\fml{D}$}
  \State{%
    \label{alg:exp:ln01}$\fml{H}\gets\emptyset$}%
  \Comment{$\fml{H}$ defined on set $S$}
  \Repeat\label{alg:exp:ln02}%
  \State{%
    \label{alg:exp:ln03} $(\outc,\mbf{r})\gets\SAT(\fml{H})$}
  \If{\label{alg:exp:ln04} $\outc=\True$}
  \State{\label{alg:exp:ln05}${A}\gets\{s_i\in{S}\,|\,r_i=1\}$}%
  \State{\label{alg:exp:ln06}${C}\gets\{s_i\in{S}\,|\,r_i=0\}$}%
  \If{\label{alg:exp:ln07}\NOT~$\cpath(\fml{D},{C})$}
  \State{\label{alg:exp:ln08}$X\gets\axp(\fml{D},{A})$}
  \State{\label{alg:exp:ln09}$\prtaxp(X)$}
  \State{\label{alg:exp:ln10}$\fml{H}\gets\fml{H}\cup\{(\lor_{s_i\in{X}}\neg{s_i})\}$}
  \Else\label{alg:exp:ln11}
  \State{\label{alg:exp:ln12}$X\gets\cxp(\fml{D},{C})$}
  \State{\label{alg:exp:ln13}$\prtcxp(X)$}
  \State{\label{alg:exp:ln14}$\fml{H}\gets\fml{H}\cup\{(\lor_{s_i\in{X}}{s_i})\}$}
  \EndIf
  \EndIf
  \Until{\label{alg:exp:ln15}$\outc=\False$}
  \EndProcedure
\end{algorithmic}

  \caption{Enumeration of AXps and CXps} \label{alg:enum}
\end{algorithm}

As indicated earlier in this section, we use a
MARCO-like~\cite{lpmms-cj16} algorithm for enumerating XPs of an XpG
(see~\autoref{alg:enum}). (An in-depth analysis of MARCO is included
in earlier work~\cite{lpmms-cj16}.)
\autoref{alg:enum} exploits hitting set duality between AXps and
CXps~\cite{inams-aiia20}, and represents the sets to hit (resp.~block)
as a set of positive (resp.~negative) clauses $\fml{H}$, defined on a
set of variables $S$.
The algorithm iteratively calls a SAT oracle on $\fml{H}$ while the
formula is satisfiable. Given a model, which splits $S$ into variables
assigned value 1 (i.e.\ set) and variables assigned value 0
(i.e.\ unset), we check if the model enables the prediction to change
(i.e.\ we check the existence of a path to a terminal node labeled 0,
with $C$ as the reference set). If no such path exists, then we
extract one AXp, using $A$ as the seed. Otherwise, we extract one CXp,
using $C$ as the seed.
The resulting XP is then used to block future assignments to the
variables in $S$ from repeating XPs.

\subsection{Enumerating CXps for DTs} \label{ssec:dtcxpenum}

The purpose of this section is to show that, if the XpG is a tree
(e.g.\ in the case of a DT), then the number of CXps is polynomial on
the size of the XpG. Furthermore, it is shown that the set of all
CXps can be computed in polynomial time.
This result has
a number of consequences, some of which are discussed
in~\autoref{sec:chkxp}.
For the concrete case of enumeration of XPs of tree XpG's, since we
can enumerate all CXps in polynomial time, then we can exploit the
well-known results of Fredman\&Khachiyan~\cite{khachiyan-jalg96} to
prove that enumeration of AXps can be obtained in quasi-polynomial
time. The key observation is that, since we can enumerate all the
CXps in polynomial time, then we can construct the associated
hypergraph, thus respecting the conditions of
Fredman\&Khachiyan's
algorithms~\cite{khachiyan-jalg96,khachiyan-dam06}.

\begin{proposition} \label{prop:allcxpdt}
  For a tree XpG, the number of CXps is polynomial on the size of the
  XpG, and can be enumerated in polynomial time.
\end{proposition}

\begin{Proof}
  To change the prediction, we must make a path to a prediction
  $c'\in\fml{K}\setminus\{c\}$ consistent. In a tree, the number of
  paths (connecting the root to a terminal) associated with a
  prediction in $c'\in\fml{K}\setminus\{c\}$ is linear on the size of
  the tree.
  Observe that each path yielding a prediction other that $c$
  contributes at most one CXp, because the consistency of the path (in
  order to predict a class other than $c$) requires that all the
  inconsistent literals be allowed to take some consistent value.
  We can thus conclude that the number of CXps is linear on the size
  of a tree XpG. 
  The algorithm for listing the CXps exploits the previous remarks,
  but takes into consideration that some paths may contribute
  candidate CXps that are supersets of others (and so not actual
  CXps); these must be filtered out.
\end{Proof}

\section{Deciding Explanation Membership} \label{sec:chkxp}

To the best of our knowledge, the problem of deciding the membership
of some literal in a prime implicant has not been studied in detail
before.
However, in the case of explainability, it is paramount to be able to
answer the query of whether a feature (and assigned value) are
included in some explanation.
This section briefly analyzes the complexity of deciding membership in
PIs. Clearly, the results for AXps/CXps track the results for PIs.
For the general case of DNFs, we prove that PI membership is hard  for
$\stwop$. For general XpG's, we show that the problem is in
$\tn{NP}$. Finally, for tree XpG's, we show that  PI/AXp/CXp
membership is in $\tn{P}$. Hence, for DTs, we can decide in polynomial
time whether some feature is included in some AXp or CXp.

\begin{proposition}
  Deciding PI membership for a DNF is hard for $\stwop$.
\end{proposition}

\begin{Proof}[Sketch]
  We reduce deciding membership in a minimal unsatisfiable subset (MUS),
  which is known to be hard for $\stwop$~\cite{liberatore-aij05}, to
  PI testing for DNFs. Let $\varphi=\{\gamma_1,\ldots,\gamma_m\}$ be
  an unsatisfiable CNF formula. We want to decide whether $\gamma_i$
  is included in some MUS.
  The reduction works as follows. First, create a DNF $\neg\varphi$,
  which is valid. Then, define a boolean function
  $\psi:\mbb{S}\to\{0,1\}$, by conjoining a selector variable $s_j$
  with each term $\neg\gamma_j$. Clearly, $\psi(1,\ldots,1)=1$, and
  $\psi(0,\ldots,0)=0$. Furthermore, it is plain that any assignment
  to the $s_i$ variables that selects any MUS of $\varphi$, will be a
  prime implicant of $\psi$, and vice-versa. Hence, $\gamma_j$ is in
  some MUS of $\varphi$ iff $s_j$ is in some prime implicant of
  $\psi$.
\end{Proof}

\begin{proposition}
  Deciding PI/AXp/CXp membership for an XpG $\fml{D}$ is in
  $\tn{NP}$.
\end{proposition}

\begin{Proof}
  To show that PI membership is in $\tn{NP}$, we consider a concrete
  variable $s_i\in{S}$. Moreover, we guess an assignment to 
  the variables in $S\setminus\{s_i\}$, say $\mbf{u}\in\mbb{S}$, such
  that $u_i=1$.
  To decide the membership of $s_i=1$ in the PI represented by
  $\mbf{u}$, we proceed as follows:
  \begin{enumerate}[nosep]
  \item Check that given the assignment $\mbf{u}$,
    $\sigma_{\fml{D}}(\mbf{u})=1$. This is done in polynomial time by
    running \autoref{alg:cpath} (and failing to reach a terminal node
    with label 0).
  \item The next step is to pick each $u_j=1$ (one of which is $u_i$),
    change its value to 0, and check that
    $\sigma_{\fml{D}}(\mbf{u})=0$. This is again done by running 
    \autoref{alg:cpath} (at most $m$ times). This way we establish
    subset  minimality. 
  \end{enumerate}
  Overall, we can check in polynomial time that $\mbf{u}$ represents a
  prime implicant containing $u_i$. Hence, the membership decision
  problem is in $\tn{NP}$.
\end{Proof}  

\begin{proposition}
  Deciding PI/AXp/CXp membership for a tree XpG is in $\tn{P}$.
\end{proposition}

\begin{Proof}
  From~\autoref{prop:allcxpdt}, we know that enumeration of all CXps
  can be achieved in polynomial time. Hence, we can simply run the
  algorithm outlined in the proof of~\autoref{prop:allcxpdt}, list all
  the features that occur in CXps, and decide whether a given feature
  is included in that list.
  For AXps the membership problem is also in $\tn{P}$, simply by
  taking~\autoref{prop:dualxp} into account. For PIs, the results
  match those of AXps.
\end{Proof}

\section{Related Work} \label{sec:relw}

Our work can be related with recent work on bayesian classifiers and
decision
graphs~\cite{darwiche-ijcai18,darwiche-aaai19,darwiche-ecai20}, but
also languages from the knowledge compilation (KC)
map~\cite{marquis-kr20}. In addition, we build on the recent results
on the interpretability and the need for explainability of
DTs~\cite{iims-corr20}.
The algorithms described in some of the previous
work~\cite{darwiche-ijcai18,darwiche-aaai19,darwiche-ecai20} cover
PI-explanations (and also minimum cardinality explanations, which we
do not consider), but do not consider contrastive explanations. The
focus of this earlier work is on ordered decision diagrams, and the
proposed algorithms operate on binary features. Furthermore, the
proposed algorithms are based on the compilation to some canonical
representation (referred to as an ODD). If the goal is to find a few
explanations, the algorithms described in this paper are essentially
guaranteed to scale in practice, whereas compilation to a canonical 
representation is 
less likely to scale 
(e.g.\ see \cite{msgcin-nips20}). 
Similarly, other recent work~\cite{marquis-kr20} investigates
languages from the knowledge compilation map, which consider binary
features. 
In addition, the tractable classifiers considered
in~\cite{marquis-kr20} for AXps do not intersect those studied in this 
paper.

\section{Experimental Results} \label{sec:res}
This section presents the experiments carried out to assess the
practical effectiveness
of the proposed algorithms. The assessment is performed on the
computation of abductive (AXp) and contrastive (CXp) explanations for
two case studies of DGs:
OBDDs and DTs.
The experiments consider a
selection
of datasets that are publicly available and originate from UCI Machine
Learning Repository \cite{uci}, Penn Machine Learning Benchmarks
\cite{Olson2017PMLB} and openML \cite{OpenML2013}.
These benchmarks are organized into two categories: the first category
contains binary classification datasets with fully binary features,
and counts 11 datasets;
the second category comprises binary and multidimensional
classification datasets with categorical and/or ordinal (i.e.\ integer
or real-valued) features, and counts 34 datasets.
Hence, the total number of considered datasets is 45.
The subset of the binary datasets is considered for generating OBDDs,
while the remaining selected datasets are used for learning DTs.

To learn OBDDs, we first train Decision List (DL) models on the given
binary datasets and then compile the obtained DLs into OBDDs using the
approach proposed in \cite{narodytska-fmcad19}.
DLs are learned using Orange3 \cite{JMLR:demsar13a}, the order of
rules is determined by Orange3 and the last rule is the default
rule. The compilation to OBDDs is performed using
BuDDy~\cite{LindNielsen1999BuDDyA}. 

A rule is of the form ``IF~\textsf{\small
  antecedent}~THEN~\textsf{\small prediction}'', 
where the \textsf{\small antecedent} is a conjunction of features, and
the \textsf{\small prediction} is the class variable $y$. The
\textsf{\small antecedent} of default rule  is empty.
Rules are translated into terms, and then conjoined into a Boolean
function.
\begin{example}
  Given a DL = $\{x_1 \land x_2 \to \mathbf{1}, \bar{x}_1 \land x_2 \to \mathbf{0}, \emptyset \to \mathbf{0}\}$, it represents Boolean function:
    \begin{equation}
      \begin{aligned}
        F_{G} &= x_1 \land x_2 \land y \\
        &\lor \hspace{1mm}\overline{x_1 \land x_2} \land (\bar{x}_1 \land x_2) \land \bar{y} \\
        &\lor \hspace{1mm}\overline{x_1 \land x_2} \land \overline{\bar{x}_1 \land x_2} \land \bar{y}
      \end{aligned}
    \end{equation}
    After compilation, we compute $F_{G|y=1}$ on OBDD to eliminate
    class variable $y$, therefore any path ending in $y$
    (resp. $\bar{y})$ is now a path to $\mathbf{1}$
    (resp. $\mathbf{0}$).
\end{example}

For training DTs,  we use the learning tool IAI ({\it Interpretable
IA}) \cite{bertsimas-ml17,iai}, which provides shallow DTs that are
highly accurate.
To achieve high accuracy in the DTs, the maximum depth is tuned
to 6 while the remaining parameters are kept in their default set up.
(Note that the test accuracy achieved for the trained classifiers,
both OBDDs and DTs, is always greater than 75\%).

All the proposed algorithms are implemented in Python, in the
\texttt{XpG} package\footnote{\url{https://github.com/yizza91/xpg}}.
The PySAT package \cite{imms-sat18} is used to instrument incremental
SAT oracle calls in XP enumeration (see \autoref{alg:enum}) and the
\texttt{dd} \footnote{https://github.com/tulip-control/dd} package,
implemented in 
Python and Cython, is used to integrate BuDDy, which is implemented in
C.
The experiments are performed on a MacBook Pro with a 6-Core Intel
Core~i7 2.6~GHz processor with 16~GByte RAM, running macOS Big Sur.

\setlength{\tabcolsep}{5pt}
\rowcolors{2}{gray!10}{}
\newcommand{\lpr}{(}
\newcommand{\rpr}{)}

\begin{table*}[ht]
\centering
\resizebox{\textwidth}{!}{
  \begin{tabular}{l>{\lpr}S[table-format=4.0,table-space-text-pre=\lpr]S[table-format=3.0,table-space-text-post=\rpr]<{\rpr}S[table-format=2.0]c  c cccc  cccc  cccc}
\toprule[1.2pt]
\rowcolor{white}
\multirow{2}{*}{\bf Dataset} & \multicolumn{2}{c}{\multirow{2}{*}{\bf (\#F~~~~~~\#TI)}}  & \multicolumn{2}{c}{\bf OBDD} & \multicolumn{1}{c}{\bf XPs}  & \multicolumn{4}{c}{\bf AXp} & \multicolumn{4}{c}{\bf CXp}  &  \multicolumn{4}{c}{\bf Runtime}\\

  \cmidrule[0.8pt](lr{.75em}){4-5}
  \cmidrule[0.8pt](lr{.75em}){6-6}
  \cmidrule[0.8pt](lr{.75em}){7-10}
  \cmidrule[0.8pt](lr{.75em}){11-14}
  \cmidrule[0.8pt](lr{.75em}){15-18} 
\rowcolor{white}
& \multicolumn{2}{c}{}   & {\bf \#N} & {\bf \%A}  &  {\bf avg} & {\bf Mx} &  {\bf m} & {\bf avg} & {\bf \%L} &  {\bf Mx} &  {\bf m} & {\bf avg} & {\bf \%L} & {\bf Tot}  & {\bf Mx}  & {\bf m} &  {\bf avg}  \\
\toprule[1.2pt]
corral                   & 6    & 64  & 6   & 100 & 4  & 4  & 1 & 2 & 34 & 4  & 2 & 2 & 22 & 0.072 & 0.002 & 0.001 & 0.001 \\
dbworld-bodies           & 4702 & 62  & 7   & 92  & 4  & 2  & 1 & 1 & 1  & 3  & 1 & 2 & 1  & 0.072 & 0.002 & 0.001 & 0.001 \\
dbworld-bodies-stemmed   & 3721 & 62  & 6   & 84  & 3  & 3  & 1 & 1 & 1  & 4  & 1 & 2 & 1  & 0.056 & 0.001 & 0.000 & 0.001 \\
dbworld-subjects         & 242  & 63  & 14  & 84  & 5  & 2  & 1 & 1 & 2  & 5  & 2 & 4 & 1  & 0.090 & 0.003 & 0.001 & 0.001 \\
dbworld-subjects-stemmed & 229  & 63  & 18  & 84  & 6  & 3  & 1 & 1 & 2  & 5  & 2 & 4 & 1  & 0.190 & 0.007 & 0.001 & 0.003 \\
mofn\_3\_7\_10           & 10   & 251 & 21  & 98  & 11 & 33 & 1 & 5 & 34 & 33 & 3 & 7 & 23 & 1.183 & 0.022 & 0.001 & 0.005 \\
mux6                     & 6    & 64  & 9   & 100 & 5  & 4  & 1 & 2 & 51 & 4  & 3 & 3 & 24 & 0.103 & 0.004 & 0.001 & 0.002 \\
parity5+5                & 10   & 222 & 71  & 80  & 8  & 11 & 1 & 2 & 59 & 15 & 5 & 6 & 14 & 1.237 & 0.015 & 0.003 & 0.006 \\
spect                    & 22   & 93  & 284 & 87  & 11 & 24 & 1 & 4 & 22 & 36 & 1 & 7 & 14 & 1.726 & 0.074 & 0.007 & 0.019 \\
threeOf9                 & 9    & 205 & 33  & 95  & 8  & 16 & 1 & 3 & 39 & 18 & 3 & 5 & 21 & 0.921 & 0.017 & 0.002 & 0.004 \\
xd6                      & 9    & 325 & 11  & 100 & 7  & 18 & 1 & 4 & 34 & 27 & 3 & 3 & 18 & 0.647 & 0.010 & 0.001 & 0.002 \\
\bottomrule[1.2pt]
\end{tabular}
}
\caption{Listing all XPs (AXp's and CXp's) for OBDDs. 
Columns {\bf \#F} and {\bf \#TI} report, respectively, the number of features, 
and the number of tested instances, in the dataset. 
(Note that for a dataset containing more than 1000 instances,  30\% of its instances, 
randomly selected, are used to be explained. Moreover, duplicate rows in  the datasets are filtered.) 
Column {\bf XPs}  reports  the average number of total explanations 
(AXp's and CXp's). 
Sub-Columns {\bf \#N} and {\bf \%A} show, respectively, total number of 
nodes and test accuracy of an OBDD.  
Sub-columns {\bf Mx},  {\bf m} and {\bf avg} of column {\bf AXp} 
(resp.,  {\bf CXp} ) show, respectively, the maximum, minimum and average 
number of explanations. The average length of an explanation  (AXp/CXp) is given 
as {\bf \%L}.
Sub-columns  {\bf Tot}, {\bf Mx},  {\bf m} and {\bf avg}  of column {\bf RunTime} 
reports, respectively, the total, maximal, minimal and average time in second 
to list all the explanations for all tested instances. \\ 
} 
\label{tab:obdd}
\end{table*}

%


\begin{table*}[h!]
\centering
\resizebox{\textwidth}{!}{
  \begin{tabular}{l>{\lpr}S[table-format=3.0,table-space-text-pre=\lpr]S[table-format=4.0,table-space-text-post=\rpr]<{\rpr}cS[table-format=2.0]c  c cccc  cccc  cccc}
\toprule[1.2pt]
\rowcolor{white}

\multirow{2}{*}{\bf Dataset} & \multicolumn{2}{c}{\multirow{2}{*}{\bf (\#F~~~~~~~\#TI)}}  & \multicolumn{3}{c}{\bf DT} & \multicolumn{1}{c}{\bf XPs}  & \multicolumn{4}{c}{\bf AXp} & \multicolumn{4}{c}{\bf CXp}  &  \multicolumn{4}{c}{\bf Runtime}\\
  \cmidrule[0.8pt](lr{.75em}){4-6}
  \cmidrule[0.8pt](lr{.75em}){7-7}
  \cmidrule[0.8pt](lr{.75em}){8-11}
  \cmidrule[0.8pt](lr{.75em}){12-15}
  \cmidrule[0.8pt](lr{.75em}){16-19} 
\rowcolor{white}
& \multicolumn{2}{c}{}   & {\bf D}  & {\bf \#N} & {\bf \%A}  &  {\bf avg} & {\bf Mx} &  {\bf m} & {\bf avg} & {\bf \%L} &  {\bf Mx} &  {\bf m} & {\bf avg} & {\bf \%L} & {\bf Tot}  & {\bf Mx}  & {\bf m} &  {\bf avg}  \\
\toprule[1.2pt]

adult & 12 & 1766 & 6 & 83 & 78 & 8 & 11 & 1 & 2 & 41 & 12 & 2 & 5 & 13 & 5.76 & 0.010 & 0.001 & 0.003 \\
agaricus-lepiota & 22 & 2437 & 6 & 37 & 100 & 6 & 6 & 1 & 3 & 17 & 7 & 2 & 4 & 7 & 5.30 & 0.006 & 0.001 & 0.002 \\
anneal & 38 & 886 & 6 & 29 & 99 & 9 & 8 & 1 & 3 & 14 & 10 & 2 & 6 & 5 & 4.02 & 0.015 & 0.002 & 0.005 \\
bank & 19 & 10837 & 6 & 113 & 88 & 18 & 38 & 1 & 9 & 33 & 21 & 4 & 9 & 12 & 87.11 & 0.032 & 0.002 & 0.008 \\
cancer & 9 & 449 & 6 & 37 & 87 & 7 & 8 & 1 & 3 & 39 & 7 & 2 & 4 & 21 & 1.12 & 0.006 & 0.001 & 0.003 \\
car & 6 & 519 & 6 & 43 & 96 & 4 & 4 & 1 & 2 & 39 & 6 & 1 & 2 & 24 & 0.71 & 0.004 & 0.001 & 0.001 \\
chess & 36 & 959 & 6 & 33 & 97 & 7 & 10 & 1 & 3 & 12 & 10 & 1 & 5 & 5 & 2.99 & 0.012 & 0.001 & 0.003 \\
churn & 20 & 1500 & 6 & 21 & 75 & 2 & 1 & 1 & 1 & 5 & 1 & 1 & 1 & 5 & 1.04 & 0.002 & 0.001 & 0.001 \\
colic & 22 & 357 & 6 & 55 & 81 & 11 & 18 & 1 & 5 & 23 & 10 & 3 & 6 & 8 & 1.31 & 0.011 & 0.001 & 0.004 \\
collins & 23 & 485 & 6 & 29 & 75 & 4 & 1 & 1 & 1 & 11 & 4 & 1 & 3 & 5 & 0.58 & 0.002 & 0.001 & 0.001 \\
dermatology & 34 & 366 & 6 & 33 & 90 & 7 & 6 & 1 & 2 & 14 & 11 & 1 & 5 & 4 & 0.97 & 0.007 & 0.001 & 0.003 \\
divorce & 54 & 150 & 5 & 15 & 90 & 6 & 8 & 1 & 3 & 7 & 4 & 1 & 3 & 3 & 0.73 & 0.010 & 0.002 & 0.005 \\
dna & 180 & 901 & 6 & 61 & 90 & 10 & 28 & 1 & 4 & 3 & 12 & 2 & 5 & 2 & 32.15 & 0.097 & 0.010 & 0.036 \\
hayes-roth & 4 & 84 & 6 & 23 & 78 & 3 & 3 & 1 & 1 & 54 & 3 & 1 & 2 & 27 & 0.06 & 0.001 & 0.000 & 0.001 \\
hepatitis & 19 & 155 & 5 & 17 & 77 & 6 & 10 & 1 & 3 & 18 & 5 & 2 & 3 & 10 & 0.29 & 0.004 & 0.001 & 0.002 \\
house-votes-84 & 16 & 298 & 6 & 49 & 91 & 9 & 30 & 1 & 5 & 25 & 10 & 2 & 4 & 13 & 0.93 & 0.016 & 0.001 & 0.003 \\
iris & 4 & 149 & 5 & 23 & 90 & 5 & 3 & 1 & 2 & 58 & 4 & 2 & 3 & 39 & 0.16 & 0.003 & 0.001 & 0.001 \\
irish & 5 & 470 & 4 & 13 & 97 & 3 & 2 & 1 & 1 & 33 & 2 & 1 & 2 & 23 & 0.27 & 0.001 & 0.000 & 0.001 \\
kr-vs-kp & 36 & 959 & 6 & 49 & 96 & 7 & 23 & 1 & 4 & 12 & 8 & 2 & 4 & 5 & 3.03 & 0.014 & 0.001 & 0.003 \\
lymphography & 18 & 148 & 6 & 61 & 76 & 11 & 15 & 1 & 5 & 28 & 12 & 3 & 6 & 10 & 0.54 & 0.009 & 0.001 & 0.004 \\
molecular-biology-promoters & 58 & 106 & 6 & 17 & 86 & 4 & 6 & 1 & 2 & 6 & 5 & 1 & 2 & 3 & 0.43 & 0.008 & 0.003 & 0.004 \\
monk1 & 6 & 124 & 4 & 17 & 100 & 3 & 2 & 1 & 1 & 38 & 3 & 1 & 2 & 18 & 0.11 & 0.002 & 0.000 & 0.001 \\
monk2 & 6 & 169 & 6 & 67 & 82 & 6 & 7 & 1 & 2 & 65 & 9 & 2 & 5 & 23 & 0.31 & 0.005 & 0.001 & 0.002 \\
monk3 & 6 & 122 & 6 & 35 & 80 & 4 & 6 & 1 & 2 & 45 & 4 & 2 & 3 & 23 & 0.15 & 0.004 & 0.001 & 0.001 \\
mouse & 5 & 57 & 3 & 9 & 83 & 3 & 4 & 1 & 1 & 41 & 3 & 2 & 2 & 25 & 0.04 & 0.001 & 0.000 & 0.001 \\
mushroom & 22 & 2438 & 6 & 39 & 100 & 6 & 5 & 1 & 2 & 18 & 7 & 2 & 4 & 7 & 5.43 & 0.007 & 0.001 & 0.002 \\
new-thyroid & 5 & 215 & 3 & 11 & 95 & 4 & 2 & 1 & 1 & 54 & 3 & 2 & 3 & 21 & 0.12 & 0.001 & 0.000 & 0.001 \\
pendigits & 16 & 3298 & 6 & 121 & 88 & 8 & 12 & 1 & 2 & 37 & 13 & 5 & 6 & 9 & 10.19 & 0.011 & 0.002 & 0.003 \\
seismic-bumps & 18 & 774 & 6 & 37 & 89 & 7 & 12 & 1 & 4 & 17 & 7 & 2 & 4 & 11 & 3.02 & 0.009 & 0.001 & 0.004 \\
shuttle & 9 & 17400 & 6 & 63 & 99 & 4 & 4 & 1 & 1 & 34 & 6 & 2 & 3 & 13 & 33.67 & 0.005 & 0.001 & 0.002 \\
soybean & 35 & 622 & 6 & 63 & 88 & 7 & 4 & 1 & 1 & 15 & 7 & 3 & 5 & 4 & 3.12 & 0.012 & 0.002 & 0.005 \\
spambase & 57 & 1262 & 6 & 63 & 75 & 10 & 22 & 1 & 3 & 11 & 15 & 3 & 7 & 3 & 6.63 & 0.019 & 0.002 & 0.005 \\
tic-tac-toe & 9 & 958 & 6 & 69 & 93 & 9 & 13 & 1 & 4 & 51 & 12 & 3 & 6 & 20 & 2.94 & 0.009 & 0.001 & 0.003 \\
zoo & 16 & 59 & 6 & 23 & 91 & 5 & 2 & 1 & 1 & 24 & 6 & 2 & 4 & 9 & 0.12 & 0.003 & 0.001 & 0.002 \\

\bottomrule[1.2pt]
\end{tabular}
}
\caption{Listing all XPs (AXp's and CXp's) for DTs.    
Sub-Columns {\bf \#D}, {\bf \#N} and {\bf \%A} report, respectively, tree's max depth, 
total number of nodes and test accuracy of a DT. 
The remaining columns hold the same meaning as described in the caption of \autoref{tab:obdd}. 
}
\label{tab:tree} 
\end{table*}

\autoref{tab:obdd} summarizes the obtained results of explaining
OBDDs.
(The table's caption also describes the meaning of each column.)
As can be observed, the maximum running time to enumerate XPs is less
than 0.074 sec for all tested XpG's in any OBDD and does not exceed
0.02 sec on average.
In terms of the number of XPs, the total number of AXps and CXps per
instance is relatively small.
Thus the overall cost of the SAT oracle calls made for XP enumeration
is negligible.
In addition, these observations apply even for large OBDDs, e.g.\
OBDD learned from the \emph{spect} dataset has 284 nodes and results
in 11 XPs on average.

Similar observations can be made with respect to explanation
enumeration for DTs, the results of which are detailed in
\autoref{tab:tree}.
Exhaustive enumeration of XPs for a XpG built from a DT takes only a
few milliseconds.
Indeed, the largest average runtime (obtained for the \emph{dna}
dataset) is 0.036 sec.
Furthermore and as can be observed, the average length {\%L} of an XP
is in general relatively small, compared to the total number of
features of the corresponding dataset.
Also, the total number of XPs per instance is on average less than 11
and never exceeds 18.

Although the DGs considered in the experiments can be viewed as
relatively small and shallow (albeit this only reflects the required
complexity given the public datasets available), the run time of the
enumerator depends essentially on solving a relatively simple CNF
formula ($\fml{H}$) which grows linearly with the number of XPs. (The
run time of the actual extractors in negligible.) This suggests that
the proposed algorithms will scale for significantly larger DGs,
characterized also by a larger total number of XPs.

\section{Conclusions} \label{sec:conc}

The paper introduces explanation graphs, which allow several classes
of graph-based classifiers to be explained with the same algorithms.
These algorithms allow for a single abductive or a single contrastive 
explanation to be computed in polynomial time, and enumeration
of explanations to be achieved with a single call to a SAT oracle per 
computed explanation. The paper also relates the evaluation of
explanation graphs with monotone functions.
In addition, the paper proves that for decision trees, computing all
contrastive explanations and deciding feature membership in some
explanation can be solved in polynomial time.
The experimental results demonstrate the practical effectiveness of
the ideas proposed in the paper.

\paragraph{Acknowledgments.} This work was supported by the AI
Interdisciplinary Institute ANITI, funded by the French program
``Investing for the Future -- PIA3'' under Grant agreement
no.\ ANR-19-PI3A-0004, and by the H2020-ICT38 project COALA
``Cognitive Assisted agile manufacturing for a Labor force supported
by trustworthy Artificial intelligence''.

%
\bibliographystyle{named}

\end{document}